# CardioGenAI: A Machine Learning-Based Framework for Re-Engineering Drugs for Reduced hERG Liability

**Gregory W. Kyro***
Yale University
Department of Chemistry
New Haven, CT 06511, USA
gregory.kyro@yale.edu

**Matthew T. Martin**
Pfizer Research & Development
Drug Safety Research & Development
Groton, CT 06340, USA
matthew.martin@pfizer.com

**Eric D. Watt**
Pfizer Research & Development
Drug Safety Research & Development
Groton, CT 06340, USA
eric.watt@pfizer.com

**Victor S. Batista**
Yale University
Department of Chemistry
New Haven, CT 06511, USA
victor.batista@yale.edu

## Abstract

The link between in vitro hERG ion channel inhibition and subsequent in vivo QT interval prolongation, a critical risk factor for the development of arrythmias such as Torsade de Pointes, is so well established that in vitro hERG activity alone is often sufficient to end the development of an otherwise promising drug candidate. It is therefore of tremendous interest to develop advanced methods for identifying hERG-active compounds in the early stages of drug development, as well as for proposing redesigned compounds with reduced hERG liability and preserved on-target potency. In this work, we present CardioGenAI, a machine learning-based framework for re-engineering both developmental and commercially available drugs for reduced hERG activity while preserving their pharmacological activity. The framework incorporates novel state-of-the-art discriminative models for predicting hERG channel activity, as well as activity against the voltage-gated $Na_V1.5$ and $Ca_V1.2$ channels due to their potential implications in modulating the arrhythmogenic potential induced by hERG channel blockade. We applied the complete framework to pimozide, an FDA-approved antipsychotic agent that demonstrates high affinity to the hERG channel, and generated 100 refined candidates. Remarkably, among the candidates is fluspirilene, a compound which is of the same class of drugs (diphenylmethanes) as pimozide and therefore has similar pharmacological activity, yet exhibits over 700-fold weaker binding to hERG. We envision that this method can effectively be applied to developmental compounds exhibiting hERG liabilities to provide a means of rescuing drug development programs that have stalled due to hERG-related safety concerns. Additionally, the discriminative models can also serve independently as effective components of a virtual screening pipeline. We have made all of our software open-source to facilitate integration of the CardioGenAI framework for molecular hypothesis generation into drug discovery workflows.

# 1. Introduction

There is a well-established connection between in vitro blockade of the hERG (human Ether-à-go-go-Related Gene) potassium ion channel and in vivo QT interval prolongation, where the QT interval, as recorded on electrocardiograms (ECGs), indicates the time between the start of the heart's ventricular depolarization (i.e., the rapid influx of sodium ions that renders the cell's interior less negatively charge) and the end of repolarization (i.e., the restoration of the cell's membrane potential to its resting negative state).[1] The hERG channel contributes to repolarization of the cardiac action potential by selectively allowing potassium ions to flow out of the cell following depolarization.[2] Inhibition of this channel can therefore directly disrupt cardiac repolarization, leading to prolongation of the QT interval, which consequently elevates the risk of potentially fatal arrythmias such as Torsade de Pointes (TdP).[3] As a result, the potential propensity of drug candidates to present hERG liabilities is subject to rigorous regulatory scrutiny, and the pharmaceutical industry devotes a significant amount of resources to identifying hERG liabilities during early, preclinical and clinical phases of drug development.[4]

The Comprehensive In Vitro Proarrhythmia Assay (CiPA) initiative,[5] supported by regulatory agencies including the U.S. Food and Drug Administration (FDA), established guidelines for evaluating the proarrhythmia risk of drugs that also incorporate the voltage-gated sodium ($Na_V1.5$) and calcium ($Ca_V1.2$) ion channels alongside the hERG channel due to observations that modulating $Na_V1.5$ and $Ca_V1.2$ channel activities may mitigate the arrhythmogenic potential induced by hERG channel blockade.[6-8] A well-known example of this phenomenon is the case of verapamil, a drug that blocks both hERG and $Ca_V1.2$ channels and is known to have only a small impact on the QT interval, which is hypothesized to be due to the counteracting effects of $Ca_V1.2$ blockade.[9] Additionally, $Ca_V1.2$ blockade alone is reported to be a possible mechanism underlying undesirable blood-flow dynamics.[10] It is therefore of tremendous interest to develop highly capable methods for assessing how both prospective and currently available drugs interact with each of these three cardiac ion channels.

A multitude of experimental methods exist for in vitro determination of cardiac ion channel affinity.[11-14] However, they require synthesis of the compounds to be assayed, which is relatively time-consuming and expensive compared to in silico methods. Machine learning (ML)-based methods for predicting hERG channel activity have been extensively explored, utilizing both protein structure-based and ligand-based models.[15-39] However, structure-based predictive modeling of the hERG channel has proven to be difficult due to the channel's intricate structure, its dynamic nature encompassing multiple conformations, and the possibility of unexpected interaction sites that are not apparent in conventional structural models.[40] For these reasons, ligand-based methods currently predominate. Predictive modeling for $Na_V1.5$ and $Ca_V1.2$ channel blocking is comparatively unexplored, as the amount of available data is much less compared to that for hERG. However, recent benchmarks for predicting $Na_V1.5$ and $Ca_V1.2$ channel activity have been established,[41] and increasing effort is being devoted to developing models for these channels as well.[42-45]

While ML-based discriminative models for predicting hERG channel activity have tremendous potential for applications in virtual screening, extending these capabilities to molecular generation through generative artificial intelligence (AI) can overcome the constraints of the currently available molecular libraries by enabling the direct in silico development of drugs with desired activities against cardiac ion channels. Numerous generative models have already demonstrated the ability to produce molecules with prespecified drug-like properties,[46-105] and there has also been work aimed at generating molecules with desired on-target potency.[53, 106, 107] Despite the progress, there has been comparatively less effort devoted to developing and applying generative models for off-target potency optimization. Moreover, the abundance of available datapoints with low hERG activity, as opposed to the general scarcity of datapoints with high on-target potency for a given target, suggests that generative models for off-target potency optimization can more effectively identify patterns in the relevant chemical space and therefore be more successful than those for on-target potency optimization, further motivating method development in this area of research.

In this work, we present an ML-based framework designed to re-engineer both developmental and commercially available drugs for reduced hERG liability while retaining their pharmacological activity. The method utilizes a generative model to produce molecules conditioned on the molecular scaffold and physicochemical properties of the input hERG-active molecule. The generated ensemble is filtered using deep learning models for predicting hERG, $Na_V1.5$ and $Ca_V1.2$ channel activity. A chemical space representation is then constructed from the filtered generated distribution and the input molecule, where nearby molecules exhibit similar chemical properties, thus facilitating the identification of molecules with similar pharmacological activity to the input molecule but with reduced hERG channel inhibition. This approach, while not a replacement for the expertise of medicinal chemists, is highly effective at rapid molecular hypothesis generation, proposing refined candidates that can then be investigated with more expensive computational methods and experimental techniques.

## 2. Overview of CardioGenAI Framework

The CardioGenAI framework combines generative and discriminative ML models to re-engineer hERG-active compounds for reduced hERG channel inhibition while preserving their pharmacological activity. An autoregressive transformer is trained on a dataset that we previously curated which contains approximately 5 million unique and valid SMILES strings derived from ChEMBL 33, GuacaMol v1, MOSES, and BindingDB datasets.[108-112] The model is trained autoregressively, receiving a sequence of SMILES tokens as context as well as the corresponding molecular scaffold and physicochemical properties, and iteratively predicting each subsequent token in the sequence. Once trained, this model is able to generate valid molecules conditioned on a specified molecular scaffold along with a set of physicochemical properties. For an input hERG-active compound, the generation is conditioned on the scaffold and physicochemical properties of this compound (Figure 1A). Each generated compound is subject to filtering based on activity against hERG, $Na_V1.5$ and $Ca_V1.2$ channels. Depending on the desired activity against each channel, the framework employs either classification models to include predicted non-blockers (i.e., $pIC_{50}$ value ≤ 5.0) or regression models to include compounds within a specified range of predicted $pIC_{50}$ values. Both the classification and regression models utilize the same architecture, and are trained using three feature representations of each molecule: a feature vector that is extracted from a bidirectional transformer trained on SMILES strings, a molecular fingerprint, and a graph (more details in section 3.1). For each molecule in the filtered generated ensemble and the input hERG-active molecule, a feature vector is constructed from the 209 chemical descriptors available through the RDKit Descriptors module.[113] The redundant descriptors are then removed according to pairwise mutual information calculated for every possible pair of descriptors. Cosine similarity is then calculated between the processed descriptor vector of the input molecule and the descriptor vectors of every generated molecule to identify the molecules most chemically similar to the input molecule but with desired activity against each of the cardiac ion channels (Figure 1B).

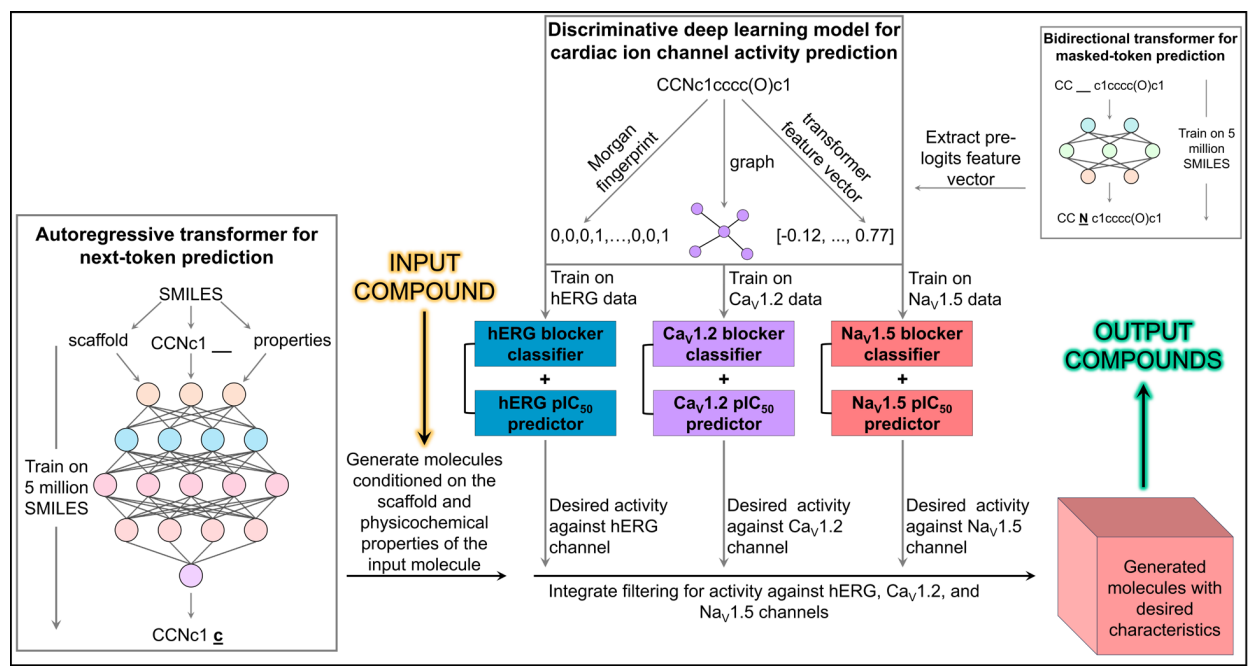

**Figure 1.** CardioGenAI framework for re-engineering hERG-active compounds. The generative transformer-based model is trained for next-token prediction, conditioned on a molecular scaffold and a set of physicochemical properties. Compounds are generated, conditioned on the scaffold and physicochemical properties of a given input compound, and the generated ensemble is filtered based on desired activity against hERG, Na$_V$1.5 and Ca$_V$1.2 channels. Cosine similarity is calculated between a descriptor vector of the input compound and that of every filtered compound to identify the most chemically similar molecules to the input compound but with desired activity against the cardiac ion channels.

## 3. Discriminative Models for Predicting Cardiac Ion Channel Activity

### 3.1. Data Featurization

For training and evaluation of hERG, Na$_V$1.5 and Ca$_V$1.2 inhibition prediction models, we utilize the training and evaluation datasets included in the benchmarks recently developed by Arab et al.[41] These benchmarks are designed to assess model generalizability, enforcing a maximum fingerprint similarity cutoff between molecules in the training and evaluation sets. Multiple published models in the field have been assessed using evaluation sets that have significant overlap with the corresponding training sets,[38, 114] undoubtedly yielding overoptimistic results with respect to the models' abilities to generalize. The compounds in the evaluation sets used in this work have a structural similarity, as determined by pairwise Tanimoto similarity between 2048-bit Morgan fingerprints, no greater than 0.70 to any compound in the corresponding training or validation sets. Compounds were sourced from the ChEMBL bioactivity database,[115-117] PubChem,[118] BindingDB,[112, 119] hERGCentral,[120] and the scientific literature.[38, 121-123] Each molecule is represented as a SMILES string which was canonicalized using RDKit, and labeled with the experimentally determined cardiac ion channel $pIC_{50}$ value. For binary classification tasks, compounds with a $pIC_{50}$ value greater than or equal to 5.0 are labeled as blockers. For hERG, $Na_V1.5$ and $Ca_V1.2$ channels, training sets contain 17 796 (78.3 %), 1 653 (74.8 %), and 641 (72.6 %) datapoints, validation sets contain 4 450 (19.6 %), 414 (18.7 %), and 161 (18.2 %) datapoints, and test sets contain 474 (2.1 %), 142 (6.4 %), and 81 (9.2 %) datapoints, respectively. For more details regarding the curation of the datasets, we refer readers to the original paper.[41]

It is important to note that different experimental protocols could contribute to differences in measured $pIC_{50}$ values for each channel due to differences in the probabilities of each channel occupying open, closed and inactivated states.[124, 125] Thus, given that the datasets used are curations of publicly available data that were obtained via different experimental protocols, variability in the experimental conditions and state probabilities may set an artificial limit on the predictive accuracy that models can achieve.

We find there to be a positive correlation (Pearson r = 0.256) between hERG $pIC_{50}$ values and the logarithm of the partition coefficient between n-octanol and water (LogP), as well as a negative correlation (Pearson r = -0.215) with topological polar surface area (TPSA) (Figure S1 in the *Supporting Information*). We also identify a relation between $pIC_{50}$ values and the presence of charged nitrogen atoms within aromatic or hydrophobic groups among the molecules exhibiting the most substantial hERG activity (Figure S2 in the *Supporting Information*).

We represent each compound as three distinct forms: a 256-dimensional feature vector that is extracted from a bidirectional transformer trained on SMILES strings, a 1024-bit Extended-Connectivity Fingerprint with a radius of 2 bonds (ECFP2) generated using the Morgan algorithm, and a graph (Figure 2). A bidirectional transformer is first trained for masked-token prediction on the same dataset used to train the autoregressive transformer, allowing it to develop an intricate internal representation of molecular structure and grasp the syntax of SMILES notation (more details in section 4.1). After this model is fully trained, it is used as a means of extracting a context-rich feature vector as a representation of a given SMILES string (more details in section 4.2). Specifically, the processed vector from the penultimate layer of the model corresponding to the start token is extracted, which contains information about the entire SMILES string that contributes to the prediction of a masked token within the sequence. This information encapsulates nuanced inter-token relationships and patterns among different molecules, rendering this feature vector a powerful representation that captures important characteristics of the molecule in a high-dimensional space. In the graph representation, nodes are atoms and edges are bonds. Each node is represented as a 14-dimenional vector of atomic features: carbon indicator, nitrogen indicator, oxygen indicator, phosphorous indicator, sulfur indicator, hydrophobicity indicator, aromaticity indicator, hydrogen bond acceptor indicator, hydrogen bond donor indicator, ring structure indicator, number of bonds to heavy atoms, number of bonds to heteroatoms, partial charge, and atomic mass. Each edge is labeled with the corresponding bond order.

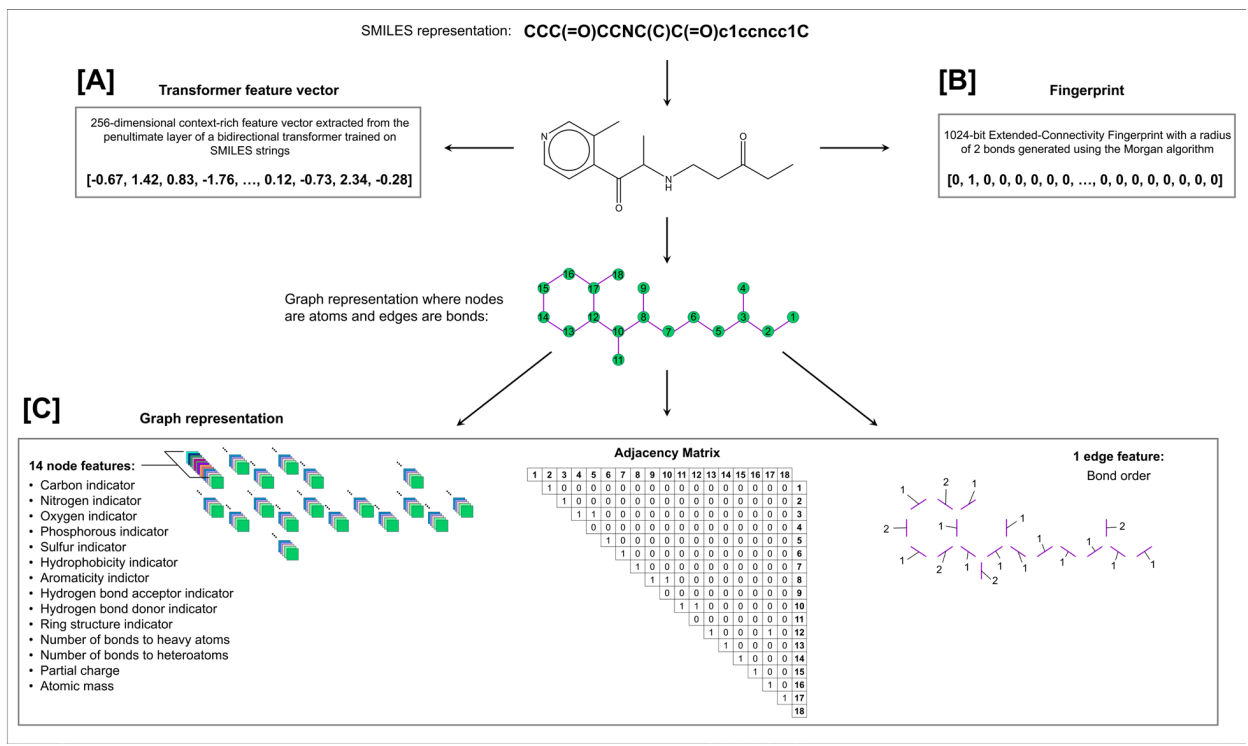

**Figure 2.** Featurization of a SMILES string for use by the cardiac ion channel activity prediction models. Each SMILES string is represented as [A] a 256-dimensional feature vector that is extracted from the penultimate layer of a bidirectional transformer trained on SMILES strings, [B] a 1024-bit Extended-Connectivity Fingerprint with a radius of 2 bonds (ECFP2) generated using the Morgan algorithm, and [C] a graph.

### 3.2. Model Architecture

The transformer-based feature vector and the ECFP2 are each processed by separate two-layer feed-forward networks (Figure 3B,C). For each of the two layers of the networks, the input vector undergoes a linear transformation followed by batch normalization. The normalized output is then passed through a ReLU activation function, followed by dropout with a rate of 50%. The graph representation is processed by a graph attention network (GAT) consisting of two GAT convolutional layers (Figure 3A). Initially, the graph is augmented with self-loops to ensure that each node's feature vector is included in its own neighborhood during feature aggregation. The first GAT layer transforms the node feature vectors through a linear operation, followed by a softmax-based attention mechanism to assign weights to the features of each node's neighbors. The output of this layer is passed through a ReLU activation function and fed to the second GAT convolutional layer which operates analogously to the first layer. After being processed by the second GAT convolutional layer, the updated node features are aggregated to form a graph-level representation using a global add pooling operation, which sums the node features across all nodes to generate a single vector that encapsulates the entire graph's information. After each of the three input feature representations has been encoded, they are concatenated to form a combined feature vector. This combined feature vector is then passed through a two-layer feed-forward network (Figure 3D). The first layer applies a linear transformation to the feature vector followed by batch normalization. The normalized output is then passed through a ReLU activation function followed by dropout with a rate of 50%. The output of this layer then undergoes a linear transformation to map it to the final output space.

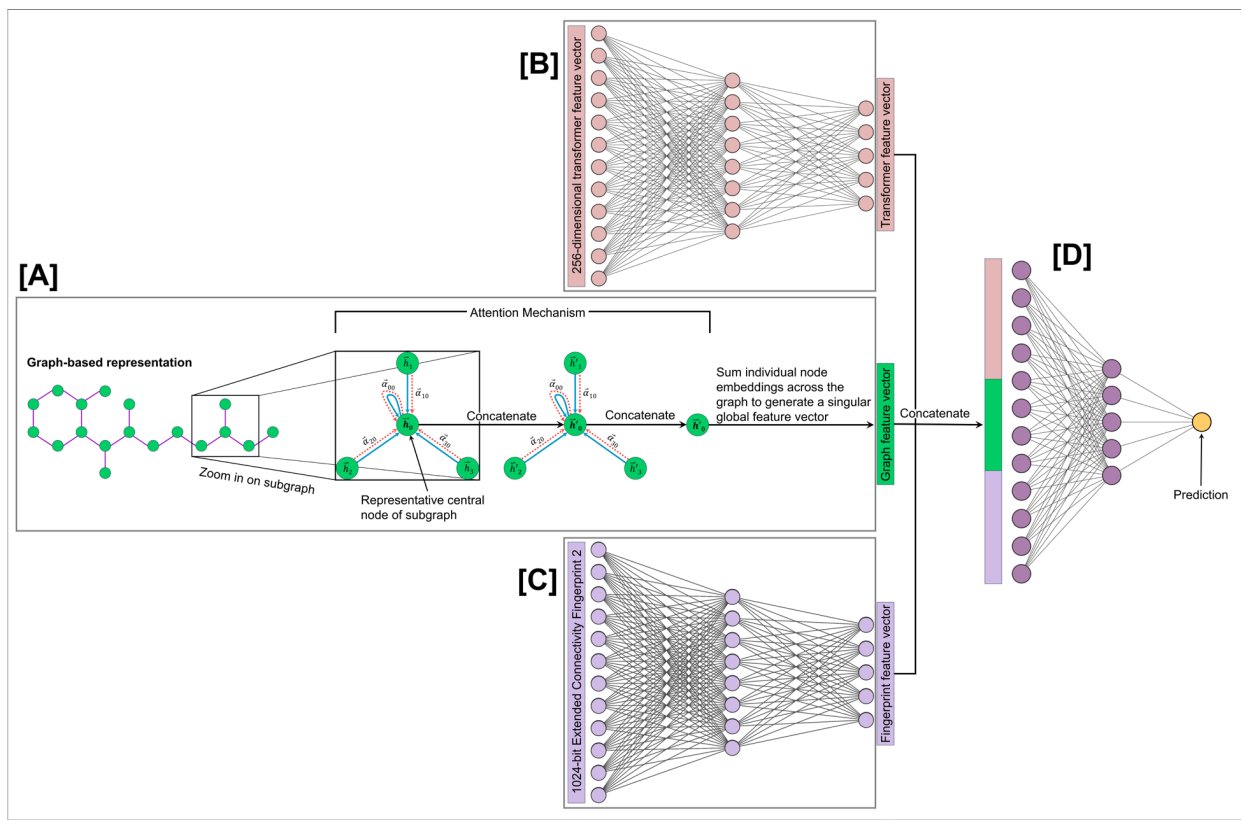

**Figure 3.** Forward pass of the cardiac ion channel activity prediction models. The graph representation of a given SMILES string is encoded by [A] a graph attention network (GAT). The [B] transformer-derived and [C] fingerprint feature vectors are encoded by feed-forward networks. These three encodings are then concatenated and passed to [D] a final feed-forward network to generate a prediction.

### 3.3. Trainings and Hyperparameters

The classification and regression models for each cardiac ion channel were trained for 200 and 100 epochs, respectively, with a batch size of 32; we trained the classification models for an additional 100 epochs because the training loss had not converged after only 100 epochs (Figure S3 of the *Supporting Information*). The AdamW optimizer, a variant of the Adam optimizer that incorporates weight decay for regularization, was used with a learning rate of $3\times10^{-4}$ and a weight decay of $1\times10^{-4}$ to optimize the models' parameters. Additionally, L1 regularization was applied with a regularization coefficient of $1\times10^{-4}$ to induce sparsity within the model parameters. We integrate a learning rate scheduler which monitors the training loss and halves the learning rate if no improvement is observed for 10 consecutive epochs. To ensure stability in training and prevent gradient explosion, gradient clipping was applied with a maximum norm of 5.0. For the classification and regression models, binary cross entropy loss and mean squared error loss were used as objective functions, respectively. The model parameters used for inference are those from the epoch with the highest validation accuracy for classification and highest validation Pearson correlation for regression. Learning curves for each of the classification and regression models are reported in Figure S3 of the *Supporting Information*.

### 3.4. Benchmarking Against Existing Models

We compare the performance of our cardiac ion channel classification models to the highest-performing models in the literature that have been evaluated with the benchmarks used in this work. Computed metrics include:

$$\text{Accuracy (AC)} = \frac{TP+TN}{TP+TN+FP+FN} \quad [1]$$

$$\text{Sensitivity (SN)} = \frac{TP}{TP+FN} \quad [2]$$

$$\text{Specificity (SP)} = \frac{TN}{TN+FP} \quad [3]$$

$$\text{F1-score (F1)} = \frac{TP}{TP+\frac{1}{2}(FP+FN)} \quad [4]$$

$$\text{Correct Classification Rate (CCR)} = \frac{SN+SP}{2} \quad [5]$$

$$\text{Matthew's Correlation Coefficient (MCC)} = \frac{TP \times TN - FP \times FN}{\sqrt{(TP+FP) \times (TP+FN) \times (TN+FP) \times (TN+FN)}} \quad [6]$$

where $TP$, $TN$, $FP$, and $FN$ represent the number of true positives, true negatives, false positives, and false negatives, respectively. We find that our model outperforms all existing models in the literature on the hERG benchmark for binary classification (Table 1). Additionally, we assess our model with different combinations of feature representations, and find that utilizing all three (i.e., transformer-based feature vector, fingerprint, and graph) achieves the best performance on the hERG benchmark (Table S4 in the *Supporting Information*).

**Table 1.** Performance of CardioGenAI for binary classification of hERG channel blockers compared to the highest-performing models in the literature on the test set presented by Arab et al.[41]

| Model | AC | SN | SP | F1 | CCR | MCC |
|---|---|---|---|---|---|---|
| CardioGenAI | **83.5** | 86.2 | **80.3** | **85.1** | **83.2** | **66.7** |
| CToxPred-hERG | 81.4 | **86.7** | 74.6 | 83.9 | 80.7 | 62.1 |
| CardioTox | 81.2 | 83.0 | 78.9 | 83.1 | 81.0 | 61.9 |
| ADMETlab 2.0 | 71.7 | 71.6 | 71.8 | 73.8 | 71.7 | 43.1 |
| ADMETsar 2.0 | 68.5 | 84.5 | 48.3 | 75.0 | 66.4 | 35.5 |
| CardPred | 56.1 | 52.7 | 60.3 | 57.0 | 56.5 | 13.0 |

[a] Compounds in the evaluation set have a structural similarity, as determined by pairwise Tanimoto similarity between 2048-bit Morgan fingerprints, no greater than 0.70 to any compound in the corresponding training or validation sets.
[b] The top value achieved for each metric is shown in bold.
[c] Accuracy (AC), sensitivity (SN), specificity (SP), F1-score (F1), correct classification rate (CCR), and Matthew's correlation coefficient (MCC) are shown.
[d] Results are shown for CToxPred-hERG,[41] CardioTox,[33] ADMETlab 2.0,[126] ADMETsar 2.0,[127] and CardPred.[21]

The improvement of our hERG predictive model over previous models justifies its use within the CardioGenAI framework as opposed to other predictive models which have already been developed.

For the $Na_V1.5$ and $Ca_V1.2$ benchmarks, only the models presented by Arab et al.[41] have been evaluated, largely owing to the fact that these benchmarks have only recently been developed and the experimental data available for these channels are scarce compared to those for hERG. We find that our models demonstrate superior performance for both cardiac ion channels (Table 2). Additionally, the area under the curve (AUC) of the receiver operating characteristic for each channel is commensurate with the accuracy that our models obtain; hERG AUC is 0.88, $Na_V1.5$ AUC is 0.89, and $Ca_V1.2$ AUC is 0.95 (Figure S5B in the *Supporting Information*).

**Table 2.** Performance of CardioGenAI for binary classification of $Na_V1.5$ and $Ca_V1.2$ blockers compared to that of the models presented by Arab et al.[41]

| Channel | Model | AC | SN | SP | F1 | CCR | MCC |
|---|---|---|---|---|---|---|---|
| $Na_V1.5$ | CardioGenAI | **89.4** | **95.9** | **75.6** | **92.5** | **85.7** | **75.1** |
|  | CToxPred-Nav | 81.7 | 85.6 | 73.3 | 86.5 | 79.5 | 58.2 |
| $Ca_V1.2$ | CardioGenAI | **91.4** | **96.2** | **82.8** | **93.5** | **89.5** | **81.0** |
|  | CToxPred-Cav | 86.4 | **96.2** | 69.0 | 90.1 | 82.6 | 70.2 |

[a] Compounds in the evaluation set have a structural similarity, as determined by pairwise Tanimoto similarity between 2048-bit Morgan fingerprints, no greater than 0.70 to any compound in the corresponding training or validation sets.
[b] The top value achieved for each metric is shown in bold.
[c] Accuracy (AC), sensitivity (SN), specificity (SP), F1-score (F1), correct classification rate (CCR), and Matthew's correlation coefficient (MCC) are shown.

We report the performance of our regression models in Figure S5C-E and Table S6 in the *Supporting Information*. We find that the Pearson correlation between true $pIC_{50}$ values and those predicted by our regression models are 0.67 for hERG, 0.60 for $Na_V1.5$, and 0.81 for $Ca_V1.2$ benchmarks (Figure S5C-E in the *Supporting Information*).

Additionally, in order to ensure that the predictive power of our models is not an artifact of spurious correlations within the data, we perform Y-randomization tests for all discriminative models and report results in Table S7 and Figure S8 of the *Supporting Information*. Furthermore, in order to provide interpretability of the regression models' predictions, we calculate the correlation between predicted $pIC_{50}$ and each property in a set of physicochemical properties for each of the three cardiac ion channels (Table S9 in the *Supporting Information*). They key findings of this analysis are as follows: predicted hERG $pIC_{50}$ correlates positively with the number of rotatable bonds (Pearson r = 0.327) and LogP (r = 0.321); predicted $Na_V1.5$ $pIC_{50}$ correlates negatively with the number of hydrogen bond donors (r = -0.593) and TPSA (r = -0.545), while correlating positively with LogP (r = 0.406); and predicted $Ca_V1.2$ $pIC_{50}$ correlates positively with the number of hydrogen bond acceptors (r = 0.621), TPSA (r = 0.581), the number of heteroatoms (r = 0.555), molecular weight (r = 0.444) and the number of rotatable bonds (r = 0.318), while correlating negatively with the number of rings (r = -0.315).

### 3.5. Application to the DrugCentral Database of FDA-Approved Drugs

To demonstrate the practical utility of our classification and regression models, we applied them to the FDA-approved drugs from the DrugCentral database, offering a real-world context for assessing cardiac ion channel inhibition.[128, 129] It is important to note that many of the compounds occur in the training set of the discriminative models. Thus, predictive ability for these compounds should not be interpreted as validation of the models' predictive ability for unseen compounds. Of the 1692 unique FDA-approved drugs, we classify 504 (29.8%) to be hERG blockers (i.e., $pIC_{50}$ value ≥ 5.0), 764 (45.2%) to be $Na_V1.5$ blockers, and 400 (23.6%) to be $Ca_V1.2$ blockers (Figure

S10A in the *Supporting Information*). A rigorous analysis of the predicted cardiac ion channel activity of the FDA-approved drugs is reported in Figure S10B of the *Supporting Information*. In addition, we report the compounds with a predicted hERG $pIC_{50}$ greater than 7.0 (i.e., more than 100-fold greater hERG inhibitory potency than the blocker threshold) in Table 3.

For the 11 FDA-approved compounds with a predicted hERG $pIC_{50}$ value greater than 7.0, the predicted $pIC_{50}$ values correlate closely with those that are experimentally determined, with notable agreement in cases where the compound is not in the training set of the model (Table 3). However, for three of the compounds, namely pimozide, astemizole, and dofetilide, the predicted hERG $pIC_{50}$ values differ from the experimentally determined values by about an order of magnitude. The experimentally determined $pIC_{50}$ values for these three compounds are among the top four highest values in the set of FDA-approved compounds, and each is greater than three standard deviations above the mean $pIC_{50}$ value in the training distribution. Because these high values are not well-represented in the training set, the model's tendency to regress toward the mean $pIC_{50}$ value likely accounts for the observed discrepancy between predicted and experimentally determined $pIC_{50}$ values for these three compounds (see Figure S5C in the *Supporting Information*).

The primary mechanism of action for three of the 11 drugs is to block the hERG channel: ibutilide,[130] dofetilide,[131] and amiodarone.[132] Another three of them function primarily as dopamine D2 receptor antagonists: pimozide,[133] droperidol,[134] and haloperidol decanoate.[135] Pimozide is reported to cause QT interval prolongation and ventricular arrhythmias due to hERG channel blockade with high specificity and affinity;[136] droperidol is reported to cause TdP due to potent hERG channel blockade;[137] haloperidol decanoate has been found to cause sudden death due to hERG channel blockade-induced QT interval prolongation.[138]

Another two of the 11 drugs function primarily as $H_1$-receptor antagonists: astemizole and terfenadine.[139, 140] Both of these drugs were withdrawn from the market due to hERG blockade-induced cardiac arrhythmias.[141, 142] Of the remaining three drugs of the 11, nintedanib is reported to cause side effects related to hERG channel blockade,[143] halofantrine is found to cause hERG blockade-induced QT interval prolongation,[144] and tolterodine is reported to cause hERG blockade-induced tachycardia and palpitations.[145] These results support the real-world application of this model to hERG activity prediction.

**Table 3.** Analysis of the FDA-approved compounds from the DrugCentral database with a predicted hERG pIC$_{50}$ greater than 7.0.

| Drug Name | Pharmacological Indication | Mechanism of Action | FDA Approval Status | Predicted hERG pIC$_{50}$ | In Training Set | Experimentally Determined hERG pIC$_{50}$ |
|---|---|---|---|---|---|---|
| Nintedanib | Idiopathic pulmonary fibrosis | Kinase inhibitor | Approved | 8.234 | yes | 8.585 |
| Ibutilide | Atrial fibrillation, atrial flutter | hERG channel blocker | Approved | 7.977 | yes | 8.000 |
| Pimozide | Tourette's disorder | Dopamine D2 receptor antagonist | Approved | 7.629 | yes | 8.520 |
| Halofantrine | Severe malaria | Forms toxic complexes with ferritoporphyrin IX | Approved | 7.588 | no | 7.398 |
| Astemizole | Allergy symptoms | H$_1$-receptor antagonist | Withdrawn due to concerns of arrhythmias | 7.562 | yes | 8.538 |
| Tolterodine | Overactive bladder | Muscarinic receptor antagonist | Approved | 7.311 | no | 7.886 |
| Droperidol | Nausea and vomiting in surgical and diagnostic procedures | Dopamine D2 receptor antagonist | Approved | 7.300 | yes | 7.495 |
| Dofetilide | Atrial fibrillation, atrial flutter | hERG channel blocker | Approved | 7.164 | yes | 8.194 |
| Haloperidol decanoate | Schizophrenia, psychotic disorders, Tourette's disorder | Dopamine D2 receptor antagonist | Approved | 7.149 | no | 6.921 |
| Amiodarone | Recurrent ventricular fibrillation, recurrent hemodynamically unstable ventricular tachycardia | hERG channel blocker | Approved | 7.127 | yes | 7.523 |
| Terfenadine | Allergic rhinitis, hay fever, allergic skin disorders | H$_1$-receptor antagonist | Withdrawn due to concerns of arrhythmias | 7.022 | yes | 7.252 |

[a] Information regarding the pharmacological indication and mechanism of action for each drug is obtained from DrugBank.[146, 147]

# 4. Transformer-Based Models

## 4.1. Data Preparation

The generative autoregressive transformer and the bidirectional transformer used for extracting features to be utilized by the discriminative models are both trained with a dataset that we previously curated by combining all of the unique and valid SMILES strings from ChEMBL 33, GuacaMol v1, MOSES, and BindingDB datasets.[108-112] The combined data set initially had a vocabulary of 196 unique tokens. To reduce the size of our vocabulary and thus improve the computational efficiency of the framework, we removed all SMILES strings containing at least one token that appeared less than 1 000 times in the combined data set; most of the SMILES strings that were excluded contain rare transition metals or isotopes. Of the remaining SMILES strings, the longest one contained 1 503 tokens, while 99.99% of the strings in the entire remaining dataset had 133 or fewer tokens. In order to reduce the block size for our transformer models, and thus further improve the computational efficiency, we removed all SMILES strings from the dataset that contain more than 133 tokens. The remaining SMILES strings were then extended, if necessary, to a length of 133 using a padding token "<pad>", and augmented with a start token "[CLS]" and an end token "[EOS]". The processed dataset contains approximately 5.5 million SMILES strings which are randomly split into training (5 262 776 entries; 95%) and validation (276 989 entries; 5%) sets. Please refer to our previous paper for complete details regarding SMILES string preprocessing.[108]

For each SMILES string, we calculate the molecular scaffold using the Murcko algorithm,[148] which identifies the core structure by removing side chains from the molecular graph, retaining the ring systems and the linkers connecting them. We also calculate ten physicochemical properties for each SMILES string: molecular weight, number of rings, number of rotatable bonds, number of hydrogen bond donors, number of hydrogen bond acceptors, TPSA, number of heteroatoms, LogP, number of stereocenters, and formal charge.

## 4.2. Model Architectures

For a given SMILES string, the autoregressive transformer considers the sequence of the SMILES string, the molecular scaffold, and the set of physicochemical properties, while the bidirectional transformer only considers the sequence. For both models, tokens in the sequence are embedded using a learnable embedding table, where each token in the vocabulary corresponds to a learnable embedding vector. The positions of the tokens in the sequence are embedded using a separate learnable embedding table, where each index in the sequence corresponds to a learnable embedding vector that allows the model to account for a given token's position in the sequence and capture sequential context within the SMILES string. For the autoregressive transformer, the set of physicochemical properties is mapped to the embedding dimension via a learnable linear transformation, and the molecular scaffold is embedded using a learnable embedding table analogous to that used for the token embeddings. For both models, all embeddings, each with an embedding dimension of 256, are summed to create a combined feature representation, and then dropout is applied with a rate of 10%.

The transformer architecture used consists of eight sequential blocks, each beginning with layer normalization to stabilize the input. This is followed by a self-attention mechanism, where query ($Q$), key ($K$), and value ($V$) vectors are computed for each input token, attention scores are derived via a scaled dot product of $Q$ and $K$ vectors, and softmax normalizes these scores to obtain weights that modulate the aggregation of $V$, effectively capturing the magnitude with which each token will attend to every other token in the sequence. The self-attention mechanism is executed multiple times in parallel through what is referred to as multi-head attention. The models used in this work employ eight attention heads, where each head uses its own set of learned linear transformations to generate $Q$, $K$, and $V$ vectors for each token in the sequence, allowing the model to simultaneously focus on different aspects of the input across the various heads. Representative attention maps for the bidirectional transformer and autoregressive transformer are reported in Figures S11 and S12 of the *Supporting Information*. The outputs of all attention heads are concatenated and passed through a learned linear transformation to generate the final output of the multi-head attention mechanism. A residual connection then merges this output with the initial block input. The resulting data tensor then undergoes another layer normalization and progresses through a two-layer feed-forward network with a 10% dropout rate and GeLU activation, before reintegration with its pre-normalized state. The final step involves another layer normalization, followed by a linear transformation that projects the data tensor onto the vocabulary space, generating a logits vector (i.e., the unnormalized log probabilities for each token in the vocabulary). When using the trained bidirectional transformer to derive feature vectors to be utilized by the discriminative models, the data tensor is extracted immediately prior to the final linear transformation, and the vector corresponding to the start token is used as the feature vector.

## 4.3. Trainings and Hyperparameters

The autoregressive transformer is trained for next-token prediction, and the bidirectional transformer is trained for masked-token prediction where each token in a given SMILES sequence has a 15% probability of being selected; of these, 80% are replaced with a mask token "<MASK>", 10% are replaced with a random token from the vocabulary, and the remaining 10% are left unchanged. Both models were trained for 100 epochs with a batch size of 512. The Sophia optimizer was used with a learning rate of $3\times10^{-4}$ and a weight decay of $1\times10^{-1}$,[149] and cross entropy loss was used as the objective function for both models. The model parameters used for inference are those from the last epoch of training. Learning curves for the autoregressive transformer and bidirectional transformer are reported in Figure S13 of the *Supporting Information*.

## 4.4. Molecular Generation

The autoregressive transformer is used to generate SMILES strings, conditioned on both a molecular scaffold and a set of physicochemical properties. To rigorously evaluate the model's ability to generate molecules with prespecified physicochemical properties, we fix one property at a time to a discrete value while the other nine properties are sampled using a random uniform distribution within ranges of acceptable values based on ADMETlab 2.0 guidelines for medicinal chemistry.[126] This procedure is performed for 500 molecules per fixed property value. For example, we generate 500 molecules conditioned on a molecular weight of 400 g/mol and another 500 conditioned on a molecular weight of 600 g/mol to assess the model's ability to generate molecules with a targeted molecular weight. We repeat this approach for each physicochemical property, and observe that the model is able to successfully generate molecular distributions that satisfy the prespecified criteria (Figure S14A-I in the *Supporting Information*). We also demonstrate the model's ability to generate molecules conditioned on multiple discrete physicochemical property values simultaneously (e.g., TPSA of 50 Å$^2$ and molecular weight of 350 g/mol), validating its utility and justifying its use within the complete CardioGenAI framework (Figure S14J in the *Supporting Information*).

# 5. Complete CardioGenAI Framework

## 5.1 High-Level Description of the Workflow

The fundamental objective of the CardioGenAI framework is to re-engineer hERG-active compounds for reduced hERG channel activity while preserving their pharmacological action. Within the framework, the autoregressive transformer first generates valid molecules conditioned on the molecular scaffold and physicochemical properties of the input hERG-active molecule, which are filtered based on desired activity against hERG, Na$_V$1.5 and Ca$_V$1.2 channels using the discriminative models. The input molecule and each filtered generated molecule are then converted into 209-dimensional chemical descriptor vectors which are refined by removing the redundant descriptors according to pairwise mutual information between every possible descriptor pair. Cosine similarity is then calculated between the descriptor vector of the input molecule and the descriptor vectors of every filtered generated molecule to identify the molecules most chemically similar to the input molecule but with desired activity against each of the cardiac ion channels.

## 5.2 Case Study: Optimizing the FDA-Approved Drug Pimozide for Reduced hERG Activity

Pimozide is an FDA-approved antipsychotic agent that is used to treat Tourette's syndrome as well as various other psychiatric disorders.[150] Its main pharmacodynamic action is to blockade dopamine D2 receptors on neurons in the central nervous system (CNS); it also has various effects on other CNS receptor systems which are not fully characterized.[133] There are many reports linking the use of pimozide to QT interval prolongation and ventricular arrythmias,[151, 152] and there are multiple reported instances of sudden, unexpected deaths of patients receiving pimozide.[153]

It was initially observed clinically that only a very low dose of pimozide is necessary to produce QT interval prolongation, suggesting that it binds to one or more cardiac potassium ion channels with high affinity.[151] Subsequent experimental validation indicated pimozide's high affinity to the hERG channel, evidenced by its potent inhibitory effect with an IC50 value of approximately 18 nM.[136]

Because of pimozide's proarrhythmic effects, it is contraindicated in patients with congenital long QT syndrome, patients with a history of cardiac arrhythmias, patients taking other drugs that prolong the QT interval, and patients with known hypokalemia (i.e., low potassium levels) or hypomagnesemia (i.e., low magnesium levels).[153] It is therefore of tremendous interest to develop safer alternatives to pimozide that minimize its hERG channel activity while retaining its therapeutic efficacy.

In this work, we apply the CardioGenAI framework to re-engineer pimozide for reduced hERG inhibition while preserving its pharmacological activity. The experimentally determined $pIC_{50}$ value of pimozide for the hERG channel is 8.520, and the value that our regression model predicts is 7.629, a difference (0.891 $pIC_{50}$) which is sufficiently small to be attributable to variance in experimental protocols used to obtain labels.[154] Our objective is to generate compounds with similar pharmacological properties, but with predicted hERG channel $pIC_{50}$ values less than 6.0. We therefore condition the molecular generation on the scaffold and physicochemical properties of pimozide, and at the generation phase, filter out molecules with a predicted hERG channel $pIC_{50}$ value greater than or equal to 6.0. This procedure is performed until 100 compounds are generated, which takes approximately one minute using an NVIDIA GeForce RTX 4050 GPU. We then compute descriptor vectors for pimozide and the filtered generated molecules, and then calculate the cosine similarity between the descriptor vector of pimozide and those of the generated molecules. In practice, many more molecules can be generated to create a molecular library for further screening.

We calculate the ten physicochemical properties for pimozide, the 100 filtered generated molecules, and the molecules in the transformer training set, and then perform principal component analysis (PCA) to construct a lower-dimensional chemical space in which we can visually compare the filtered generated molecules to pimozide in relation to the broader transformer training set. Plotting the first two PCs reveals that the filtered generated molecules are closely aligned to pimozide, indicating that our framework successfully navigates the initially vast chemical space to propose compounds with similar physicochemical characteristics to pimozide while reducing hERG activity (Figure 4A; Figure S15 in the *Supporting Information*). Additionally, the distribution of predicted $pIC_{50}$ values of the generated compounds ranges from 4.64 to 6.00 with a mean value of 5.59, indicating significant reductions in hERG activity (Figure 4B).

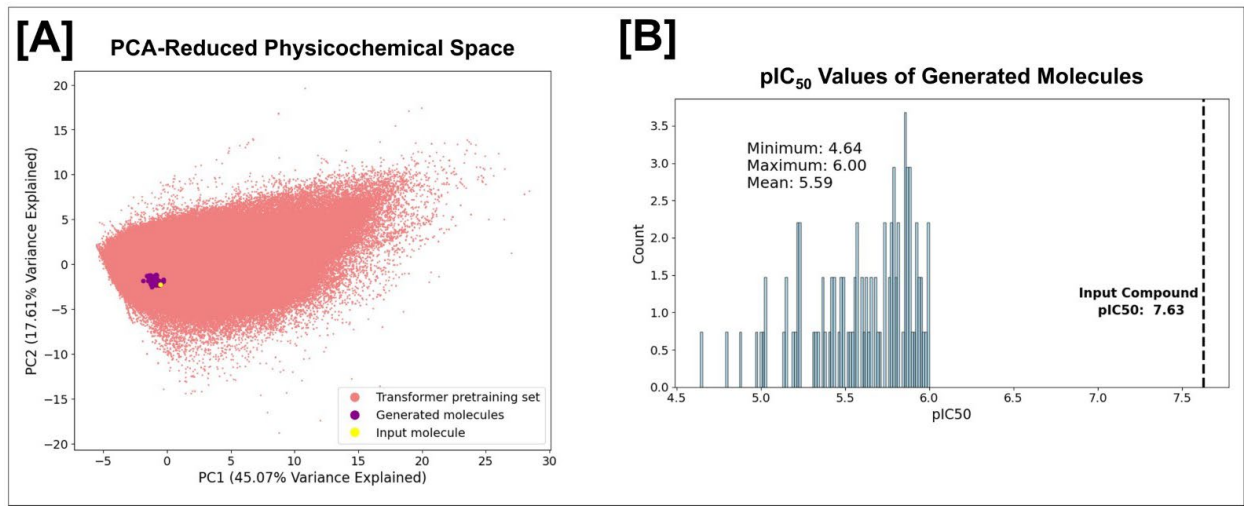

**Figure 4.** Visualization of the CardioGenAI framework applied to pimozide. The input molecule (pimozide), the 100 generated refined molecules, and the molecules in the training set for the transformer-based models (approximately 5 million datapoints), are projected into a principal component analysis (PCA)-reduced physicochemical-based space, shown in [A]. Pimozide is colored yellow, the generated refined compounds are colored purple, and the compounds in the training set of the transformer-based models are colored red. The first two principal components explain 45.07% and 17.61% of the total variance, respectively. Clearly, the CardioGenAI framework is able to identify the region of physicochemical space corresponding to compounds that are similar to pimozide, yet exhibit significantly reduced activity against the hERG channel. The density of predicted $pIC_{50}$ values against the hERG channel of the generated refined compounds as compared to that of pimozide is shown in [B]. The distribution of generated compounds exhibits a maximum predicted $pIC_{50}$ value of 6.00, with a mean of 5.59 and minimum of 4.64.

We analyze each generated refined compound with respect to all of the compounds provided in the DrugCentral Postgres v14.5 database to identify any compounds approved by either the FDA, the European Medicines Agency (EMA), or the Pharmaceuticals and Medical Devices Agency of Japan (PMDA).[128, 129] Remarkably, among the 100 filtered generated compounds is fluspirilene, a compound that belongs to the same class of drugs (diphenylmethanes) as pimozide and therefore presents a highly similar pharmacological profile.[155] Moreover, the experimental hERG channel $pIC_{50}$ value for fluspirilene is 5.638 (predicted: 5.785), as compared to 8.520 (predicted: 7.629) for pimozide (Figure 5), indicating a reduction in hERG channel affinity by over 700-fold. This case study demonstrates the ability of the CardioGenAI framework to re-engineer a hERG-active compound for reduced hERG activity while preserving its pharmacological activity. The most similar generated molecules to pimozide are reported in Table S16 of the *Supporting Information*.

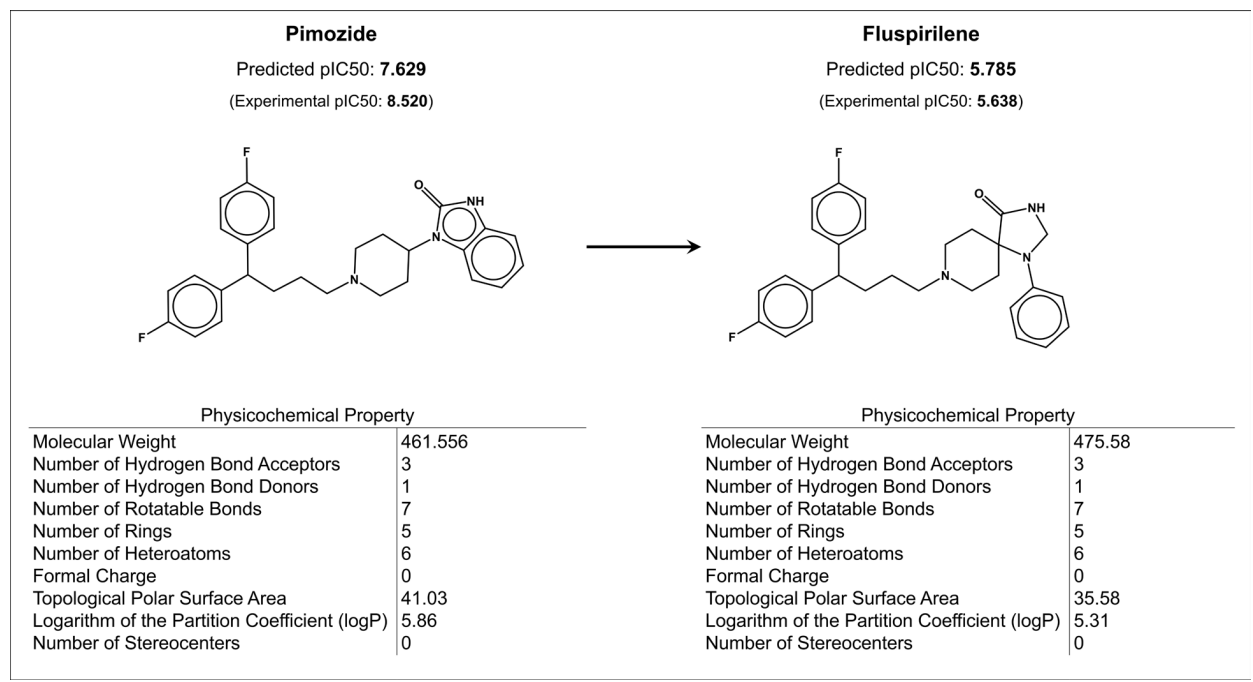

**Figure 5.** CardioGenAI framework applied to pimozide, an FDA-approved antipsychotic drug that has an experimental hERG channel $pIC_{50}$ value of 8.520 (predicted: 7.629) , and is reported to cause hERG channel blockade-induced QT interval prolongation and arrhythmias. 100 molecules are generated, and among them is fluspirilene, a compound that belongs to the same class of drugs as pimozide but exhibits significantly less hERG channel activity (experimental $pIC_{50}$ value is 5.638).

## 5.3 Additional Applications of the Complete Framework

In addition to re-engineering pimozide, we also apply the CardioGenAI framework to nintedanib, ibutilide, halofantrine, and astemizole. Collectively, including pimozide, these five compounds are those among the set of FDA-approved compounds provided by DrugCentral that have the highest predicted $pIC_{50}$ values against the hERG channel. We show that for each drug, the framework is able to successfully generate compounds with similar physicochemical profiles and with significantly reduced activity against the hERG channel (Figure S17A-J in the *Supporting Information*).

Moreover, given that modulating $Na_V1.5$ and $Ca_V1.2$ channel activities may mitigate the arrhythmogenic potential induced by hERG channel blockade,[6-8] and considering that activity against each of these two channels alone can present problems related to cardiac repolarization,[10,45] we demonstrate the ability of the framework to optimize compounds for enhanced $Na_V1.5$ and $Ca_V1.2$ profiles. Specifically, we assess the capabilities of the framework with respect to four objectives: (1) Increase the $Na_V1.5$ activity of a compound that has high hERG activity but low $Na_V1.5$ activity; (2) Increase the $Ca_V1.2$ activity of a compound that has high hERG activity but low $Ca_V1.2$ activity; (3) Decrease the $Na_V1.5$ activity of a compound that has high $Na_V1.5$ activity; (4) Decrease the $Ca_V1.2$ activity of a compound that has high $Ca_V1.2$ activity. For cases (1) and (2), we chose to re-engineer ibutilide, which has a predicted $pIC_{50}$ for hERG, $Na_V1.5$, and $Ca_V1.2$ of 7.98, 4.24 and 4.02, respectively. For case (3), we chose venetoclax, which has a predicted $Na_V1.5$ $pIC_{50}$ of 6.72. For case (4), we chose itraconazole, which inhibits $Ca_V1.2$ with a predicted $pIC_{50}$ of 9.17. The CardioGenAI framework is able to successfully improve the cardiac ion channel activity by at least one order of magnitude in each case for every generated refined compound while ensuring that the generated compounds are physicochemically similar to the respective input drug. The results for each of these four cases are presented in Figure S18 of the *Supporting Information*.

## 6. Summary

Although numerous generative models have demonstrated the ability to produce molecules with prespecified drug-like properties, as well as molecules with desired on-target potency, there has been comparatively less effort devoted to developing and applying generative models for off-target potency optimization. In this work, we present an ML-based framework for re-engineering hERG-active compounds for reduced hERG channel activity while preserving their pharmacological activity. The method utilizes an autoregressive transformer-based generative model to produce molecules conditioned on the molecular scaffold and set of physicochemical properties of the input molecule. The generated ensemble is filtered based on hERG, $Na_V1.5$ and $Ca_V1.2$ activity using discriminative deep learning models. A chemical space representation is then constructed from the filtered generated distribution and the input molecule, where nearby molecules exhibit similar chemical properties, thus facilitating the identification of molecules with similar pharmacological activity to the input molecule but with reduced hERG liability. We applied the framework to pimozide, an FDA-approved antipsychotic agent that demonstrates high affinity to the hERG channel, and generated a compound of the same class of drugs that has a significantly lower hERG $pIC_{50}$ value as indicated by both predicted and experimental values. Moreover, the abilities of the discriminative models to accurately predict $pIC_{50}$ values for the $Na_V1.5$ and $Ca_V1.2$ channels, despite the scarcity of available data, suggest that the framework can be effectively extended to encompass drug-target interactions across various targets for which data is limited.

## 7. Customizing the CardioGenAI Framework for Company-Specific Industrial Applications

Pharmaceutical companies have begun to leverage generative AI-based methods for specific tasks within the earlier stages of drug discovery pipelines.[156] In order to facilitate integration of CardioGenAI into drug discovery workflows, all of the software is entirely open-source and the framework is designed to be easily customizable. Companies can therefore incorporate desired functionality, and retrain all of the models on their in-house data. It is expected that large pharmaceutical companies will significantly benefit from retraining the models, given that their proprietary data is likely more comprehensive and subject to significantly less experimental variance than the publicly available datasets used to initially train the models.

With respect to the incorporation of additional functionality into the framework, CardioGenAI is designed such that predictive models can easily be integrated into the post-generation filtering phase along with the cardiac ion channel activity prediction models. For instance, a team of medicinal chemists will likely adhere to a well-defined list of synthesis-related criteria, and a model fit to these criteria can easily be incorporated. The objective of such a model could be to identify compounds that are produced via specific synthetic pathways, or to predict a company-specific synthetic accessibility score. In theory, any predictive model can be integrated into the framework (e.g., for predicting on-target activity, solubility, metabolic stability, bioavailability, etc.).

Because synthesizability is arguably the most important characteristics of a proposed compound, additional steps can be taken, aside from incorporating more models, to ensure that the proposed compounds are in accordance with a company's specific synthesis capabilities. For instance, the dataset used to train the generative autoregressive transformer could be curated to contain only compounds that a company deems sufficiently synthesizable, thereby biasing the generative component of the framework to only propose compounds that are akin to those that satisfy these synthesizability standards. Additionally, rather than defining the chemical space based on RDKit descriptors to identify molecules that are physicochemically similar to the input molecule, the space can be designed such that nearby molecules are produced via similar synthetic pathways.

## 8. Software Details and Availability

The transformer-based models and the feed-forward networks in the discriminative models were built using PyTorch.[157] The parameters of the transformer-based models were optimized using the Sophia optimizer.[149] The GAT components of the discriminative models were built using PyTorch Geometric.[158] The hyperparameters of the discriminative models were optimized using Optuna.[159] The hyperparameters that were optimized include: batch size, learning rate, weight decay, the number of GAT attention heads used in the graph model, the output dimension of the GAT mechanism used in the graph model, and the dropout rate applied to the fully connected components of the complete architecture. SMILES canonicalization, as well as the calculations of physicochemical properties and molecular scaffolds were performed using RDKit.[113] Scikit-learn was used to calculate pairwise mutual information between chemical features and cosine similarity between descriptor vectors, as well as to perform PCA.[160]

All of our software is available as open source at https://github.com/gregory-kyro/CardioGenAI. Users can easily run the complete CardioGenAI framework (Figure 6), perform inference with the discriminative models (Figure 7), and reproduce the figures in this manuscript. Additionally, we provide all of the data we use, as well as the parameters for each of our trained models.

### Running the CardioGenAI Framework

To optimize a cardiotoxic compound with CardioGenAI, utilize the `optimize_cardiotoxic_drug` function from the `Optimization_Framework` module:

```python
from src.Optimization_Framework import optimize_cardiotoxic_drug

optimize_cardiotoxic_drug(input_smiles,
                         herg_activity,
                         nav_activity,
                         cav_activity,
                         n_generations,
                         device)
```

- `input_smiles (str)`: The input SMILES string of the compound that you seek to optimize for reduced cardiac ion channel activity.
- `herg_activity (tuple or str)`: hERG activity for which to filter. If the entry is a string, it must be either 'blockers' or 'non-blockers'. If it is a tuple, it must indicate a range of activity values.
- `nav_activity (tuple or str)`: NaV1.5 activity for which to filter. If the entry is a string, it must be either 'blockers' or 'non-blockers'. If it is a tuple, it must indicate a range of activity values.
- `cav_activity (tuple or str)`: CaV1.2 activity for which to filter. If the entry is a string, it must be either 'blockers' or 'non-blockers'. If it is a tuple, it must indicate a range of activity values.
- `n_generations (int)`: The number of optimized drug candidates to generate. Default is 100.
- `device (str)`: The device to use for the optimization. Must be either 'gpu' or 'cpu'. Default is 'gpu'.

**Figure 6.** Python function to run the complete CardioGenAI framework.

### Performing Inference with the Discriminative Models

To predict activity against the hERG, NaV1.5 and CaV1.2 channels, utilize the `predict_cardiac_ion_channel_activity` function from the `Discriminator` module:

```python
from src.Discriminator import predict_cardiac_ion_channel_activity

predict_cardiac_ion_channel_activity(input_data,
                                     prediction_type,
                                     predict_hERG,
                                     predict_Nav,
                                     predict_Cav,
                                     device)
```

- `input_data (str or list)`: The input data for which the discriminative models will process. If the entry is a string, it must be either a SMILES string or a path to a prepared h5 file. If it is a list, it must be a list of SMILES strings.
- `prediction_type (str)`: Either 'regression' or 'classification'. Default is 'regression'.
- `predict_hERG (bool)`: Whether to predict hERG activity. Default is True.
- `predict_Nav (bool)`: Whether to predict NaV1.5 activity. Default is False.
- `predict_Cav (bool)`: Whether to predict CaV1.2 activity. Default is False.
- `device (str)`: The device to use for the inference computations. Must be either 'gpu' or 'cpu'. Default is 'gpu'.

**Figure 7.** Python function to perform inference with the discriminative models.

# Supporting Information

The Supporting Information is available free of charge at:
github.com/gregory-kyro/CardioGenAI/blob/main/results/CardioGenAI_SI_JCIM_clean_v1.pdf
- Details regarding the datasets used, model trainings, additional analyses of the models, and the refined drug candidates

# Author Information

Corresponding Authors:

Gregory W. Kyro
- Phone: (516) 413-1143
- Email: gregory.kyro@yale.edu

Victor S. Batista
- Phone: (203) 432-6672
- Email: victor.batista@yale.edu

Present Address: Department of Chemistry, Yale University, New Haven, CT 06511-8499

Author Contributions:

GWK, EDW, MTM, VSB conceived the idea; GWK, EDW, MTM designed research; GWK developed software; GWK performed research; GWK, EDW, MTM analyzed data; GWK, EDW, MTM wrote the paper; VSB provided feedback on the paper. All authors have given approval to the final version of the manuscript.

Funding Sources:

- National Science Foundation Graduate Research Fellowship: Grant DGE-2139841
- National Science Foundation Engines Development Award – Advancing Quantum Technologies (CT): Award Number 2302908
- CCI Phase I – National Science Foundation Center for Quantum Dynamics on Modular Quantum Devices (CQD-MQD): Award Number 2124511
- Yale University: seed funding

## Acknowledgments

We acknowledge financial support from the National Science Foundation Graduate Research Fellowship under Grant DGE-2139841 [GWK], from the National Science Foundation Engines Development Award: Advancing Quantum Technologies (CT) under Award Number 2302908 [VSB], and from the CCI Phase I: National Science Foundation Center for Quantum Dynamics on Modular Quantum Devices (CQD-MQD) under Award Number 2124511 [VSB]. Additionally, we acknowledge seed funding from Yale University. We also acknowledge high-performance computer time from the National Energy Research Scientific Computing Center and from the Yale University Faculty of Arts and Sciences High Performance Computing Center.

We also thank Todd A. Wisialowski, Peter J. Kilfoil, and Nathaniel Woody for their valuable comments and expert insights regarding the manuscript.

## Abbreviations

hERG, human Ether-à-go-go-Related Gene; ECGs, electrocardiograms; TdP, Torsade de Pointes; CiPA, The Comprehensive In Vitro Proarrhythmia Assay; FDA, U.S. Food and Drug Administration; $Na_V1.5$, voltage-gated sodium ion channel subtype 1.5; $Ca_V1.2$, voltage-gated calcium ion channel subtype 1.2; ML, machine learning; AI, artificial intelligence; LogP, logarithm of the partition coefficient between n-octanol and water; TPSA, topological polar surface area; ECFP2, Extended-Connectivity Fingerprint with a radius of 2 bonds; GAT, graph attention network; AC, accuracy; SN, sensitivity; SP, specificity; CCR, correct classification rate; MCC, Matthew's correlation coefficient; AUC, area under the curve; $Q$, query vector; $K$, key vector; $V$, value vector; CNS, central nervous system; PCA, principal component analysis.